\documentclass{article}

\PassOptionsToPackage{numbers, sort&compress}{natbib}
\usepackage{graphicx}
\usepackage{hyperref}
\usepackage{multirow}
\usepackage[final]{neurips_2021}
\usepackage{comment}
\usepackage{amsmath}

\usepackage[utf8]{inputenc} 
\usepackage[T1]{fontenc}    
\usepackage{hyperref}       
\usepackage{url}            
\usepackage{booktabs}       
\usepackage{amsfonts}       
\usepackage{nicefrac}       
\usepackage{microtype}      
\usepackage{color}         

\title{Hierarchical Learning in Euclidean Neural Networks}

\author{%
  Joshua A. Rackers \\
  Center for Computing Research\\
  Sandia National Laboratories\\
  Albuquerque, NM 87110 \\
  \texttt{jracker@sandia.gov} \\
 \And
  Pranav Rao\\
  Center for Computing Research\\
  Sandia National Laboratories\\
  Albuquerque, NM 87110 \\
   \\
  Institute for Condensed Matter Theory\\
  University of Illinois, Urbana-Champaign\\
  Urbana, IL 61801\\
  \texttt{pvrao2@illinois.edu} \\
}

\begin{document}

\maketitle

\begin{abstract}
Equivariant machine learning methods have shown wide success at 3D learning applications in recent years. These models explicitly build in the reflection, translation and rotation symmetries of Euclidean space and have facilitated large advances in accuracy and data efficiency for a range of applications in the physical sciences. An outstanding question for equivariant models is why they achieve such larger-than-expected advances in these applications. To probe this question, we examine the role of higher order (non-scalar) features in Euclidean Neural Networks (\texttt{e3nn}). We focus on the previously studied application of \texttt{e3nn} to the problem of electron density prediction, which allows for a variety of non-scalar outputs, and examine  whether the nature of the output (scalar $l=0$, vector $l=1$, or higher order $l>1$) is relevant to the effectiveness of non-scalar hidden features in the network. Further, we examine the behavior of non-scalar features throughout training, finding a natural hierarchy of features by $l$, reminiscent of a multipole expansion. We aim for our work to ultimately inform design principles and choices of domain applications for {\tt e3nn} networks.
\end{abstract}

\section{Introduction}
Euclidean Neural Networks are graph-based neural network models that explicitly build in the symmetries of the Euclidean group $E(3)$ in three dimensions\cite{smidt2021euclidean,geiger2022e3nn}. These models are equivariant to rotations and translations: the features of the network explicitly transform under the action of the Euclidean group. There can be scalar ($F_{l_h=0}$), vector ($F_{l_h=1}$), and higher order ($F_{l_h>1}$) hidden features at each node $i$ of the graph network, where $l_h$ represents the $(2l +1)$-dimensional irreducible representation of the rotation group $SO(3)$\footnote{We use the schematic notation of Miller et. al. \cite{miller2020relevance}. In practice the features are represented by spherical harmonics \cite{e3nn}.} :
\begin{equation}
    {\bf F}_i = \begin{pmatrix}
    F_{l_h=0}\\
    F_{l_h=1}\\
    F_{l_h=2}\\
    .\\
    .\\
    \end{pmatrix}.
\end{equation}
As a result, there is a natural data efficiency provided by equivariance; symmetry is built into the model explicitly so one does not have to resort to data augmentation (expanding training data to include symmetry-transformed samples). This makes Euclidean networks a natural choice for modeling problems in the physical and biological sciences, from crystalline materials\cite{chen2021direct} to molecules\cite{batzner20223} (and more), where systems of study are sensitive to rotations and translations of real space. 

In practice, the advantage offered by \texttt{e3nn} is more dramatic than expected, going beyond the efficiency gain from avoiding augmentation\cite{geiger2022e3nn}. Seen across multiple problem domains\cite{batzner20223,rackers2022cracking}, this advantage offers great promise for modeling environments where generating training data presents a scaling problem\cite{frey2022neural}. Still, the nature of these observed effects remains elusive; a recent overview of the \texttt{e3nn} framework bluntly noted: "Unfortunately we have no theoretical explanation for this change [of slope in the learning curve]." In this work we aim to provide observations on Euclidean Neural Networks that will help illuminate the cause of this unexpected increase in data efficiency. In particular, we seek to understand further the role of non-scalar features in \texttt{e3nn}.

Initial progress has been made by systematically establishing the advantage of non-scalar features over invariant models and scalar-only models. Previous works by Miller et. al. \cite{miller2020relevance}, Brandstetter et. al. \cite{brandstetter2021geometric} both establish the advantage of $l=1$ features over invariant models through ablation studies. Further, it was posited that equivariant models, with non-scalar hidden features, are particularly suited to learning non-scalar outputs such as vectors\cite{brandstetter2021geometric,batzner20223,miller2020relevance}.  There is a solid intuition for the first observation, namely that the equivariant graph convolution\cite{brandstetter2021geometric},
\begin{equation}
\label{eq:groupconv}
    {\bf F}'_i \sim \sum\limits_{j \in \mathcal{N}(i)} \sum\limits_l \sum\limits_{m=-l}^l {\bf F}_j \otimes \left[ R(||{\bf x}_j - {\bf x}_i||) Y_{lm}\left(\frac{{\bf x}_j - {\bf x}_i}{||{\bf x}_j -{\bf x}_i||}\right)\right],
\end{equation} 
is able to utilize both distances between neighboring nodes $|||{\bf x}_j - {\bf x}_i||$ as well as relative directional information through the spherical harmonics (the convolution is taken over neighbors $\mathcal{N}(i)$ of node $i$ and $R(x)$ is a multi-layer perceptron). For example, in the $l=1$ case, Eq.~\ref{eq:groupconv} has access to angles between nodes as well as distances. On the other hand, invariant message-passing graph networks in the literature\cite{schutt2018schnet,klicpera2020directional} have been restricted to learning only on distances between nodes. 

However, while establishing the benefit of $l>0$ models, this leaves open the nuance of what particular $l$ is necessary for a desired application, and the motivation for that specific choice. In practice, Batzner et al. \cite{batzner20223} as well as Rackers et. al \cite{rackers2022cracking} observe for independent tasks that models up to $l=2$ (but no higher) provide increasing benefits in a model's learning curve. Here, we focus on tackling the following questions:

\begin{enumerate}
    \item \textit{Do equivariant models have an advantage over invariant models specifically for learning non-scalar outputs?} 

    \item \textit{For a given task, is there an $l^\mathrm{max}_h$ for the hidden layers beyond which efficiency gains saturate? Does this change with the nature of the task?}

    \item \textit{Is there any internal structure to features in equivariant models?}
\end{enumerate}

In this work, we will address these questions in the context of electron density prediction for water clusters. The electron density prediction task is instructive because the data efficiency advantages of non-scalar features in \texttt{e3nn} have already been established for this task\cite{rackers2022cracking} and the representation of the electron density contains higher-order spherical harmonic outputs. 

We propose three sets of experiments that address the above questions. First, we examine the effect of non-scalar features in the network on non-scalar outputs. Second, we study how the maximum angular momentum of the output, $l^\mathrm{max}_o$, affects the optimal angular momentum channel in the hidden layers, $l^\mathrm{max}_h$. Finally, we look directly at how the learned features of an \texttt{e3nn} electron density model evolve over training. These experiments will help answer the questions we have laid out and shed light on the bigger question of the unexplained advantage of equivariance.

\section{Methods}
For the electron density learning task we seek to predict the coefficients of the density represented in a density fitting basis\cite{rackers2022cracking,fabrizio2019electron}):
\begin{equation}
\label{eq:densityfit}
\rho({\bf r}) = \sum\limits_{\mu=0}^{N_\mathrm{atoms}}\sum\limits_{\nu=0}^{N_\mathrm{basis}}\sum\limits_{l=0}^{l_\mathrm{max}} \sum\limits_{m=-l}^{l} C^{\mu\nu}_{lm} Y_{lm} e^{-\alpha_{ikl}({\bf r - r_\mu})^2}.
\end{equation}
The density is represented by a linear combination of atom-centered Gaussian basis functions, where $Y_{lm}$ are spherical harmonics, $\alpha^{\mu\nu}_l$ control the widths of each Gaussian function, and $C^{\mu\nu}_{lm}$ are coefficients for each basis function. The first two indices correspond to atom and basis indices, while the lower indices are rotational degrees of freedom associated with the spherical harmonics. The machine learning task is to learn the coefficients $C^{\mu\nu}_{lm}$, which themselves can be decomposed corresponding to each spherical harmonic: for each basis function corresponding to $l$, there are $2l+1$ associated coefficients. 

As such, the task of learning molecular electron densities is a natural one for studying how equivariant models work when tasked with predicting rotational outputs. For our purposes we focus on water molecules, and work with the density fitting basis (also referred to as ``auxiliary" basis)  \texttt{def2-universal-jfit} \cite{weigend2006accurate}, which has outputs 
\begin{equation}
\label{eq:def2}
l_{out} =12\times 0e+5\times 1o+4\times 2e+2\times 3o+1\times 4e.
\end{equation} 
The notation denotes multiplicity $\times$ $l$ parity. The fact that there are outputs transforming in $l>0$ representations for this density learning task make it a natural setting to investigate the novel advantages offered by non-scalar hidden features in \texttt{e3nn}. The quality of the fit is assessed by the density difference\cite{rackers2022cracking},
\begin{equation}
    \epsilon_\mathrm{total}(\%)= 100 \times \frac{\int d{\bf r} |\rho_\mathrm{QM}({\bf r}) - \rho_\mathrm{ML}({\bf r})|} {\int d{\bf r} \rho_\mathrm{QM}({\bf r})}.
\end{equation}
Above, $\rho_{QM}({\bf r})$ represents the density computed from Eq.~\ref{eq:densityfit} with training coefficients $(C^\mathrm{QM})^ {\mu\nu}_{lm}$ from traditional electronic structure methods. In our case, we use training data from Density Functional Theory (DFT), and used the PBE0 density functional for its accuracy on electron densities \cite{medvedev2017density,rackers2022cracking} (we refer to quantum coefficients as $C^\mathrm{DFT}$ for the rest of this work). The machine-learned electron density $\rho^\mathrm{ML}$ is computed the same way, but with the output coefficients from our {\tt e3nn} model. The integrals are evaluated on a grid of spacing $0.5$ \AA. 

\subsection{Euclidean neural network model}

\begin{figure}[t]
    \centering
    \includegraphics[scale=.5]{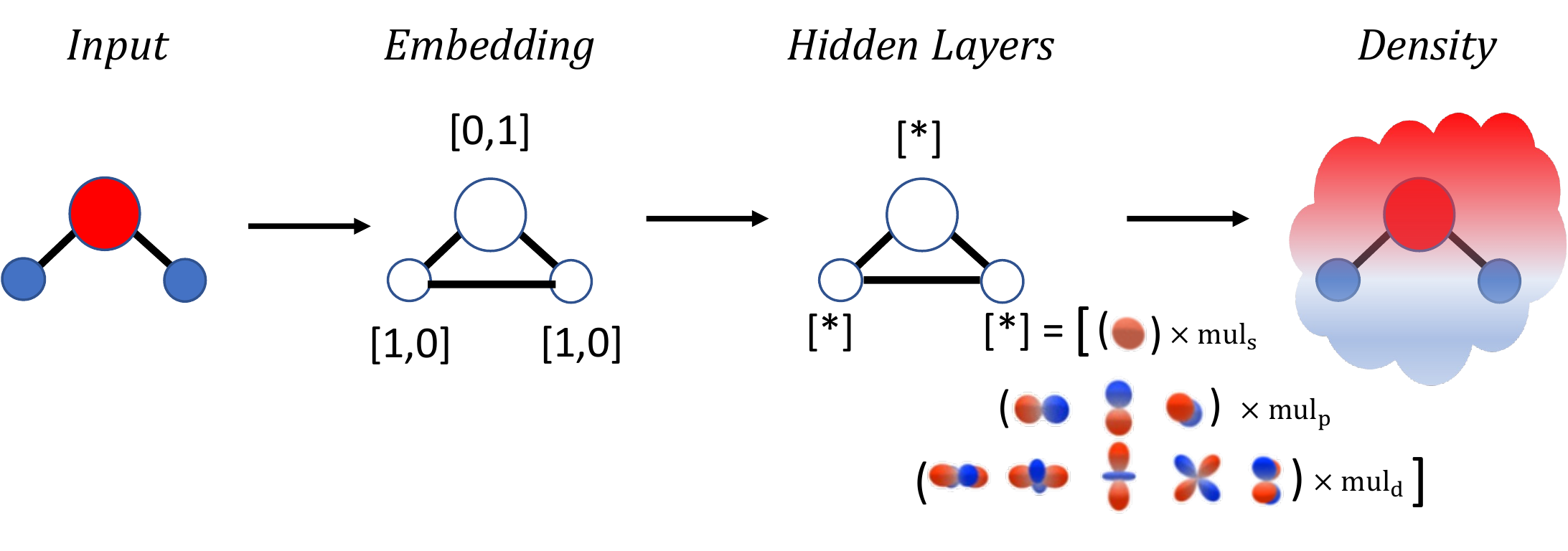}
    \caption{\texttt{e3nn} network scheme for density prediction.}
    \label{fig:network}
\end{figure}

A high level description of the models trained in this work is shown in Figure~\ref{fig:network}. Atomic types are one-hot encoded $H \rightarrow (1,0)$ and $O \rightarrow (0,1)$ and fed into the network as input along with atomic coordinates in the form of nuclear positions. The input is then passed to hidden layers consisting of an equivariant convolution operation, followed by a gated non-linearity. For each of these layers, we specify the desired irreducible representations (irreps) of the desired output. Finally, the output of the hidden layers is passed to an output layer which outputs the irreps of our electron density representation specified in Eq.~\ref{eq:def2}. This model is implemented in \texttt{e3nn} as \texttt{gate\_points\_2021}. A more complete description of this model can be found in the supplementary information of reference \cite{rackers2022cracking}.

\subsection{Datasets}
\label{sec:datasets}
We generated three separate types of datasets for the experiments below.

\begin{itemize}
    \item [] {Standard Dataset: We performed DFT calculations on 576 structures of eight water molecule clusters with the PBE0 functional and aug-cc-pvtz basis set. Then we performed density fitting with the \texttt{def2-universal-jfit} auxiliary basis set to obtain the target coefficients.\footnote{Standard data is available at \href{https://zenodo.org/record/5563139}{https://zenodo.org/record/5563139}}}.

    \item [] {Truncated Dataset: Starting from the Standard Dataset, we systematically removed high angular momentum basis functions. The full \texttt{def2-universal-jfit} auxiliary basis has $l^\mathrm{max}_o=4$, so we created truncated versions with $l^\mathrm{max}_0\in\lbrace 3,2,1,0\rbrace$. For example,
    \begin{equation}
    l^\mathrm{max}_\mathrm{o} = 3 \Longleftrightarrow 12\times 0e+5\times 1o+4\times 2e+2\times 3o 
    \end{equation}
    }

    \item [] {Scaled Dataset: The norms of the coefficients of the Standard Dataset progressively decrease with increasing $l$. To remove this decay behavior, we scaled every $l>0$ output channel of the Standard Dataset to have the same standard deviation as the $l=0$ output channel.}
\end{itemize}
\subsection{Training details}
The models in this work are trained for 500 epochs with a train-test split percentage of 75/25. 
\begin{table}
\centering
\begin{tabular}{ ||c|p{8cm}|c||}
 \hline
 \multicolumn{3}{||c||}{{{\tt e3nn} Hidden layer configurations}} \\
 \hline
 & {Hidden feature irreps} & {Parameters}\\
 \hline
$l_h = 0$   & \small${\ 525 \times 0e +525 \times 0o }$ & 2,358,430\\
\hline
 $l_h = 1$ &  \small $420 \times 0e + 420 \times 0o + 35 \times 1e + 35 \times 1o$ & 2,180,606\\ 
 \hline
$l_h = 2$ &  \small $315 \times 0e + 315 \times 0o + 35 \times 1e + 35 \times 1o + 21 \times 2e + 21 \times 2o$ & 1,741,458\\
 \hline
 $l_h = 3$ &  \small $210 \times 0e + 210 \times 0o + 35 \times 1e + 35 \times 1o + 21 \times 2e + 21 \times 2o + 15 \times 3e + 15 \times 3o $ & 1,339,006\\ 
 \hline
 $l_h = 4$ &  \small $210 \times 0e + 210 \times 0o + 35 \times 1e + 35 \times 1o + 21 \times 2e + 21 \times 2o + 15 \times 3e + 15 \times 3o + 11 \times 4e + 11 \times 4o$& 922,236 \\
 \hline
\end{tabular} 
\vspace{.1cm}
\caption{Hidden layer configurations for Experiments 1 \& 2.}
\label{tab:hiddenlayers}
\end{table}
\section{Results}
\subsection{Experiment 1: error by $l$ channel}
In order to tackle {\it Question 1} posed in the introduction - whether non-scalar outputs are learned better by including non-scalar features in an equivariant model - we examined the effect of varying the maximum angular momentum features in the hidden layers on individual $l_o$ output channels. We introduce another metric to assess the quality of our density models:
\begin{equation}
\label{eq:epsperchannel}
\epsilon_l = 100\times \frac{\int d{\bf r} |\rho^l_\mathrm{DFT}({\bf r}) - \rho^l_\mathrm{ML}({\bf r})|} {\int d{\bf r} \rho_\mathrm{DFT}({\bf r})}.
\end{equation}
This "per-channel" error quantifies the deviation of the density coming only from particular $l,m$ coefficients in Eq.~\ref{eq:densityfit}. Explicitly,
\begin{equation}
\rho^l({\bf r}) = \sum\limits_{\mu=0}^{N_\mathrm{atoms}}\sum\limits_{\nu=0}^{N_\mathrm{basis}} \sum\limits_{m=-l}^{l} C^{\mu\nu}_{lm} Y_{lm} e^{-\alpha_{ikl}({\bf r - r_i})^2}.
\end{equation}
As an example, $\epsilon_{l=0}$ corresponds to the deviation of the scalar coefficients $(C^\mathrm{ML})^\nu$ from $(C^\mathrm{DFT})^\nu$ (there are twelve such coefficients for each atom, owing to the auxiliary basis Ref.~\ref{eq:def2}). We normalize by the DFT density to keep the same relative scale for each $\epsilon_l$ and the total $\epsilon_\mathrm{total}$, but note that the sum $\sum_l \epsilon_l > \epsilon_\mathrm{total}$ (these are different quantities).

As mentioned earlier, the expectation from previous work\cite{brandstetter2021geometric,batzner20223,miller2020relevance} is generally that equivariant methods are uniquely suited to deal with non-scalar outputs. To investigate this further in the context of electron densities, we consider this per-channel error (Eq.~\ref{eq:epsperchannel}) for a variety of hidden layer configurations with varying $l^\mathrm{max}_h$.
\begin{figure}
    \centering
    \includegraphics[scale=.75]{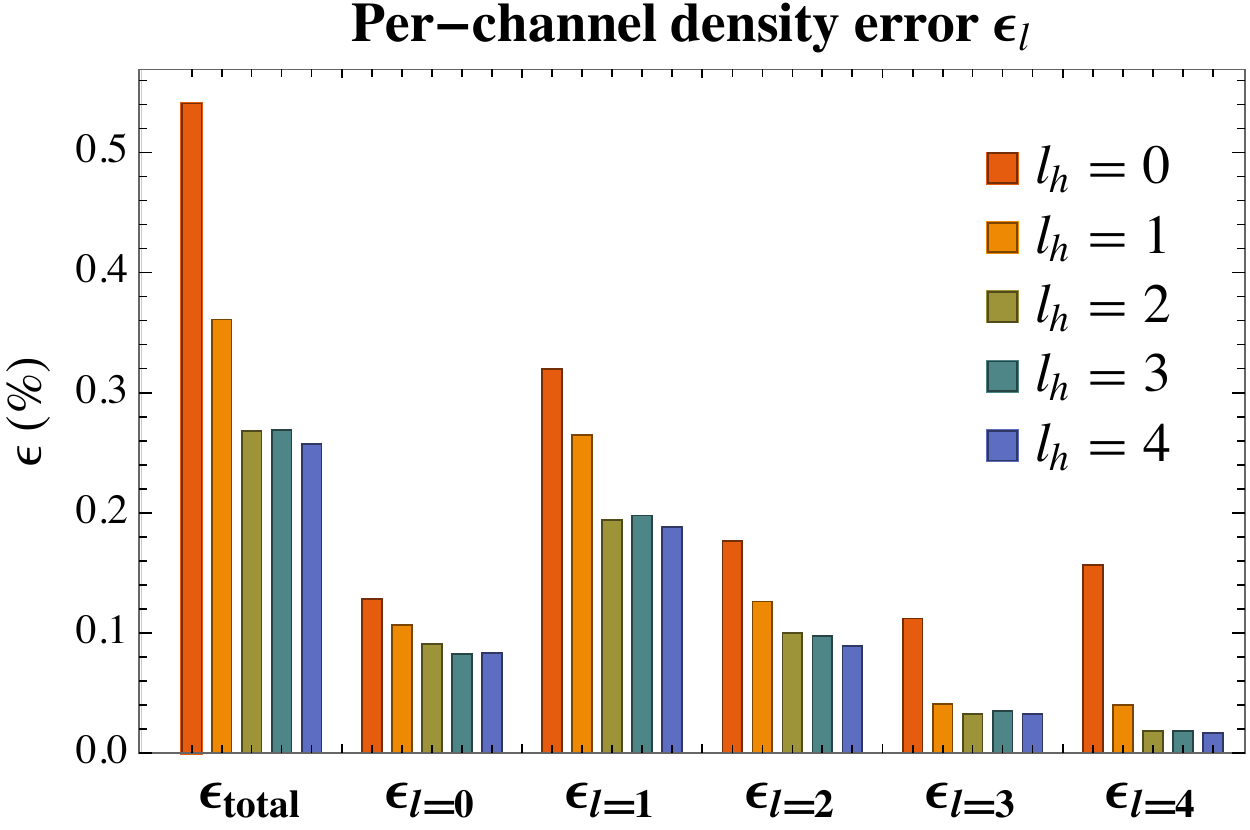}
    \caption{Density error by output angular momentum channel.}
   \label{fig:epsperchannel} 
\end{figure}
Figure.~\ref{fig:epsperchannel}, presents the results of this experiment. The notation $l_h = 0, 1, 2, 3, 4$ denotes the different hidden layer configurations used (see Table.~\ref{tab:hiddenlayers}). We start with a fully scalar model $l_h = 0$ and then successively move a fixed amount of features from the scalar channel ($N_s = 105 \times 0e + 105 \times 0o$) into the next highest $l$ channel with $N_l = N_s/(2l + 1)$. 

As seen previously\cite{rackers2022cracking}, the total density error $\epsilon_\mathrm{total}$ decreases as we increase $l_h$. However, this improvement in accuracy as we increase $l_h$ does not correspond to increases in any particular channel, but rather an overall improvement in $\epsilon_l$ for each $l$. This suggests that the addition of angular features does not confer a particular ability to learn a particular type of output, but rather an overall benefit. In other words, we see no preferential learning of non-scalar quantities as we increase the number of non-scalar hidden features in our network. 

\subsection{Experiment 2: $l^\mathrm{max}_h$ for truncated data}

Our second experiment addresses {\it Question 2} in the introduction: Is there an $l^\mathrm{max}_h$ for the hidden layers beyond which efficiency gains saturate, and can this change based on the nature of the task. 

Previous work on learning the electron density with Euclidean Neural Networks showed that learning curves improve up to $l_h^\mathrm{max}=2$, but then saturate.\footnote{See Fig.~1 of \cite{rackers2022cracking}} Here, Figure.~\ref{fig:epsperchannel} shows the same behavior. Diminishing performance returns are achieved after $l_h^\mathrm{max}=2$; this result is robust to differences in training in the present work and Rackers et. al \cite{rackers2022cracking}.
\begin{figure}[h]
    \centering
    \includegraphics[scale=.75]{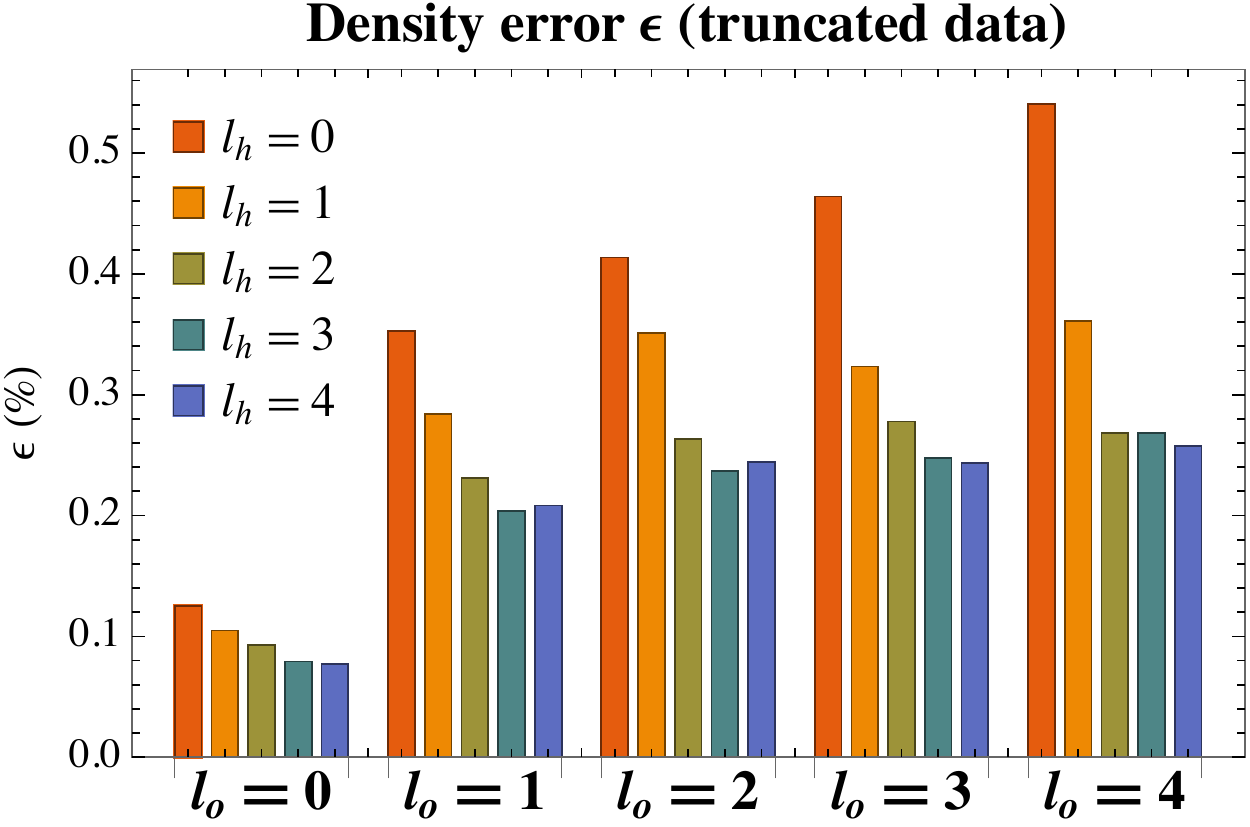}
    \caption{Density error on various truncated datasets. The advantage for each roughly saturates at $l^\mathrm{max}_h = 2$. }
  \label{fig:truncateddata}
\end{figure}
We now consider deviations from this baseline model by truncating higher order $l$ functions from the auxiliary basis (Eq.~\ref{eq:def2}), which, in its standard form, contains outputs up to $l_\mathrm{o} = 4$. We truncated this basis to create four new reference datasets (see Sec.~\ref{sec:datasets}). 

We then trained new models on each of these new truncated reference outputs. Figure.~\ref{fig:truncateddata} shows that for each of these truncated datasets we still see $l^\mathrm{max}_h = 2$. The presence or absence of non-scalar outputs does not change the optimal design of our {\tt e3nn} network. Notably, this appears to be true even for the models trained on the $l_o=0$ and $l_o=1$ datasets.

This result suggests that it is the structure of the \textit{input} graph, not the output type, that is largely responsible for dictating what higher order features are necessary. We note that Batzner et. al. \cite{batzner20223} found the same behavior of $l^\mathrm{max}_h = 2$ when considering the rather different task of learning on energies, still on water, while exploring possible hidden features up to $l=3$. This is a striking observation, and suggests that the input graph structure and geometry is ultimately responsible for what types of non-scalar features are necessary and effective in building an equivariant model.

We offer two routes for further exploration. First, considering the graph convolution Eq.~\ref{eq:groupconv}, we noted that non-scalar features allow for the inclusion of relative positional information between atoms, which was a basis\cite{brandstetter2021geometric} for considering e.g. $l=1$ networks as opposed to invariant models. Higher order features, in this context $l=2$, seem to add to the possible geometric information able to be captured by the convolution. We leave a theoretical study of this for future work. On the other hand, we leave open the possibility that $l_h^\mathrm{max} = 2$ is a result of the message-passing structure of ${\tt e3nn}$, which uses two-body messages (convolutions), and for future work it would be interesting to consider the effect of multi-body messages\cite{batatia2022design,batatia2022mace}.

\subsection{Experiment 3: Feature hierarchy}

We turn our attention to {\it Question 3}, whether the features in a Euclidean Neural Network have any internal structure throughout training. For a minimal model with hidden features up to $l=2$, we examine the norm squared for each angular momentum feature channel throughout training. These norms are averaged over all nodes and multiplicities, but to account for the dimensionality of $l>0$ features we divide the norms by $2l +1$.
\begin{figure}
    \centering
    \includegraphics[scale=.55]{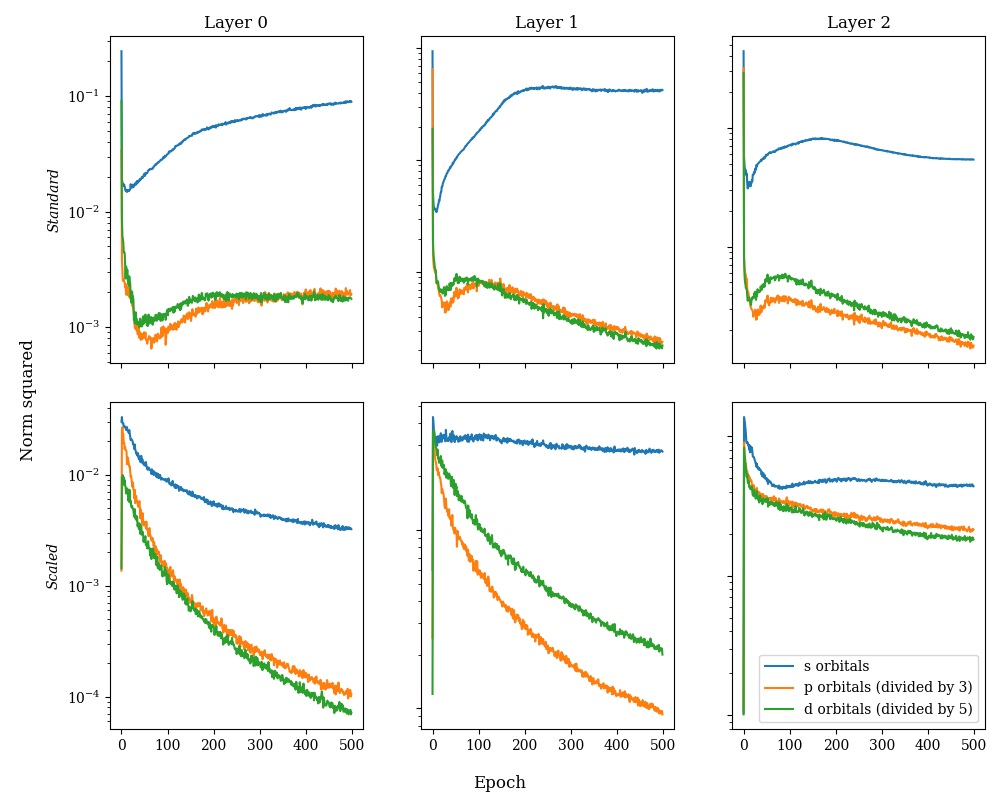}
    \caption{Features throughout training for a minimal model (hidden layers $15\times 0e + 15\times 1o  + 15 \times 2e$), trained against the standard auxiliary basis dataset as well as a "scaled" dataset. We label scalar features as $s$, vector $l=1$ features as $p$ and $l=2$ features as $d$.}
    \label{fig:features}
\end{figure}
In Figure.~\ref{fig:features} we have plotted the norm squared of each type of feature for the test evaluated at every training epoch for the standard auxiliary basis dataset as well as a ``scaled" dataset. The scaled dataset multiplies the output coefficients for each $l>0$ output channel to have the same standard deviation as the scalar $l=0$ outputs. In practice $\sigma_{l=0} \sim .8$, and we fix $\sigma^\mathrm{scaled}_{l>0} \sim .8$ which represents a large increase. 

We see in Figure.~\ref{fig:features} that scalar features dominate in importance throughout training, indicating a hierarchy between scalar and higher order features. It is interesting to note that the ordering of importance of the hidden features follows physical intuition. In physics, the most general way to approximate a point-wise function that depends on angles in Euclidean space is a multipole expansion. The most common example of a multipole expansion is to represent the electrostatic potential around a set of points as a sum of terms including progressively higher spherical harmonic contributions, generally decaying as $l$ is increased. This decaying behavior is exactly what is seen with the hidden features of our trained \texttt{e3nn} networks. 

Importantly, this behavior is seen without any physical priors of this decay behavior built into the model. An initialized \texttt{e3nn} network sees no difference between an $l=0$ feature and any other feature. In fact, this behavior is learned even when the "well-behaved" nature is taken out of the output data. In the second row of Figure.~\ref{fig:features}, even when the magnitude of the output channels are scaled to be equal (not decay with increasing $l$), the hierarchical feature learning behavior persists. This suggests that \texttt{e3nn} may be learning some physically meaningful, multipole-like internal representation of the data.

\section{Discussion}
In this work we have investigated the nature of higher order features in Euclidean Neural Networks, motivated by their remarkable but unexplained data efficiency. Our findings are useful on a practical level for the more informed design of equivariant models and also represent progress in understanding the role and effectiveness of rotational features in \texttt{e3nn}. 

It is somewhat natural to assume that equivariant models are best suited to angular outputs. This was the subject of {\it Question 1} in the introduction, which we answered for the case of electron density prediction in Experiment 1, showing no preferential learning of any particular output channel while varying the structure of the hidden features of the model. In particular, we saw that as we added higher order angular features,  the improvement in the overall test error $\epsilon_\mathrm{total}$ simply corresponded to overall improvements in the per-channel error $\epsilon_l$. As a result we see that the advantage of equivariant models is likely present for any 3D learning problem, as opposed to restricted to high angular momentum output signals like the electron density. 

It is worth pointing out that in Experiment 1, even as the dimension of the hidden representation is held constant, the actual number parameters in these models \textit{decreases} as $l_h$ increases. This is because in the higher $l_h$ networks, more of the degrees of freedom in the hidden features transform together (eg. all three components of a vector feature cannot rotate independently). This observation is remarkable; in nearly every other context, a decrease in the number of parameters would make a network less expressive. Here the added constraints of requiring the network to respect the symmetries of the Euclidean group appear to do just the opposite.

In {\it Question 2}, we further examine the relationship between the input and output structure of equivariant models. We do so by truncating the maximum $l$ in the output to $l_o^\mathrm{max}$ and considering again models for each hidden layer structure in Table.~\ref{tab:hiddenlayers}.  We find in Experiment 2 that, strikingly, regardless of the (maximum) output dimension $l_o$, the benefits of equivariance trail off above hidden layer angular momentum $l_h^\mathrm{max}$ = 2. Furthermore, we noted that the specific value of $l_h^\mathrm{max} = 2$ was seen for the different task of learning forces and energies of water \cite{batzner20223}. Together with the results of Experiment 1, we see somewhat counter intuitively that the advatanges of equivaraince over invariant models do not seem to depend on the presence of non-scalar outputs.  Rather, the ideal internal representation structure appears to be dictated by the structure of the input data itself. 

Geometrically, each water molecule has an OH bond distance and HOH bond angle that lie in a narrow range. Additionally, because of hydrogen bonding, water molecules in a cluster are not randomly distributed. There are characteristic lengths and angles that define the first and second solvation shells around a water molecule. It has been argued that the presence of vectors ($l_h=1$ features) in the hidden layers of a network allows the convolution operation to have access to angles between nodes, a three-body term, as well as distances. Our observations indicate that including up to four-body terms is essential to capturing the interactions between atoms in water clusters. In addition to geometry, we are interested in future work considering the role of message passing (specifically from two-body to many-body messages) in dictating the internal structure of \texttt{e3nn} networks. 

The observation that $l>0$ features are optimal for a range of 3D molecular learning problems is interesting in light of recent theoretical work on equivariant learning algorithms. Villar et al. show in a recent paper, "Scalars are Universal", that it is possible to construct a complete set of equivariant functions entirely with scalars.\cite{villar2021scalars} While this is mathematically true, our observations suggest that a scalar-only representation may not be the most efficient representation for all physics-based learning problems.

The clear answer to \textit{Question 3} in the introduction is: Yes, \texttt{e3nn} networks exhibit a clear, hierarchical internal structure. At a deeper level, Experiment 3 also provides a useful probe for the theoretical basis of why equivariant neural networks with $l_h>0$ exhibit advantages over invariant models. The evolution of learned hidden features in \texttt{e3nn} networks shows a distinct behavior. At the beginning of training, features from all angular momentum channels change rapidly and somewhat randomly. After this initial phase, however, a pattern emerges where the magnitude of scalar features dominates, followed by $l>0$ features. This concerted behavior across layers is a clear indicator of feature learning in \texttt{e3nn} networks. It is thought that feature learning like this can result in large gains in data efficiency over the opposing regime of kernel learning or "lazy learning" for many applications.\cite{geiger2020disentangling} While this does not conclusively prove that the degree of feature learning is behind the advantage of equivariant models with $l_h>0$ hidden features, it suggests that the feature learning behavior these models exhibit may play a part.

Machine learning has been slow to have the same impact on the physical sciences as it has on areas like computer vision or natural language processing. The simple explanation often given is that these are areas with large amounts of labeled data. However, many of the most important advances in these are areas have not been simply a function of big data. The most successful models have managed to build in some relevant symmetry of the problem. Examples include convolutions for image recognition or attention-based models for speech-to-text. This observation has led to the prediction that in the coming age of machine learning "Physics is the New Data"\cite{kalinin2022physics}. Understanding the exact role of equivariance in this paradigm is rather important in this context, and our work provides concrete observations that make steps toward doing so.

\begin{ack}
The authors thank Mario Geiger for helpful comments and suggestions. This research was supported in part by an appointment to the Universities Research Association (URA) Summer Fellowship Program at Sandia National Laboratories, sponsored by URA and administered by the Oak Ridge Institute for Science and Education.

This article has been authored by an employee of National Technology \& Engineering Solutions of Sandia, LLC under Contract No. DE-NA0003525 with the U.S. Department of Energy (DOE). The employee owns all right, title and interest in and to the article and is solely responsible for its contents. The United States Government retains and the publisher, by accepting the article for publication, acknowledges that the United States Government retains a non-exclusive, paid-up, irrevocable, world-wide license to publish or reproduce the published form of this article or allow others to do so, for United States Government purposes. The DOE will provide public access to these results of federally sponsored research in accordance with the DOE Public Access Plan \url{https://www.energy.gov/downloads/doe-public-access-plan}.

This paper describes objective technical results and analysis. Any subjective views or opinions that might be expressed in the paper do not necessarily represent the views of the U.S. Department of Energy or the United States Government.
\end{ack}

\small 
\bibliographystyle{plainnat}
\bibliography{refs}

\end{document}